\documentclass[a4paper,fleqn]{cas-dc}

\usepackage[authoryear]{natbib}
\usepackage{bbm}
\usepackage{textcomp}
\usepackage{subfigure}
\usepackage{multirow}
\usepackage{makecell}
\usepackage{arydshln} 
\usepackage{xcolor}
\usepackage[ruled,lined]{algorithm2e}

\def\tsc#1{\csdef{#1}{\textsc{\lowercase{#1}}\xspace}}
\tsc{WGM}
\tsc{QE}

\begin{document}
\let\WriteBookmarks\relax
\def\floatpagepagefraction{1}
\def\textpagefraction{.001}
\let\printorcid\relax

\shorttitle{}    

\shortauthors{Tianling Liu et al.}

\title [mode = title]{Context-driven Missing-Modality Learning for Robust Medical Diagnosis with Image-Tabular Data}  



%
















\author[1]{Tianling Liu}
\ead{liu_dling@tju.edu.cn}



\author[2]{Lequan Yu
\corref{correspondingauthor}}
\ead{lqyu@hku.hk}

\author[3,4]{Tong Han}
\ead{mrbold@163.com}

\author[1,5]{Liang Wan
\corref{correspondingauthor}}
\ead{lwan@tju.edu.cn}

\cortext[correspondingauthor]{Corresponding author: Lequan Yu, Liang Wan.}

\address[1]{College of Intelligence and Computing, Tianjin University, Tianjin 300350, China.}


\address[2]{Department of Statistics and Actuarial Science, School of Computing and Data Science, The University of Hong Kong, Hong Kong.}

\address[3]{Department of Radiology, Tianjin Huanhu Hospital, Tianjin 300350, China.}

\address[4]{Tianjin Key Laboratory of Cerebral Vascular and Neurodegenerative Diseases, Tianjin 300350, China.}

\address[5]{Medical School of Tianjin University, Tianjin 300072, China.}


\begin{abstract}
While multimodal data integrating diverse imaging and clinical tabular records is crucial for accurate medical diagnosis, the arbitrary absence of specific modalities is prevalent in clinical practice, severely degrading the performance of multimodal models. Existing methods either discard missing modalities, leading to information loss, or struggle to synthesize them without capturing complex inter-modal dependencies. 
To address these limitations, we propose a novel Context-driven Missing-Modality Learning (CMML) framework, which sequentially performs modality synthesis and semantic alignment to achieve robust diagnosis under arbitrary missing conditions. 
Specifically, we design a Cascade Residual Transformer-based Autoencoder (CRTA) that leverages learnable context tokens acting as dataset-level semantic prior to capture inter-modal dependencies and synthesize key missing representations. These representations are further enriched by modality-specific memory banks. 
To resolve the discrepancy between original available and synthesized representations, we transform the learned context tokens into instance-adaptive semantic references by infusing multimodal representations from the CRTA's outputs. This reference guides the alignment of heterogeneous modality representations into a unified space, where class-aware contrastive refinement is finally applied to explore discriminative diagnostic cues.
Extensive evaluations on skin lesion (Derm7pt), ocular disease (ODIR), and meningioma (MEN) datasets demonstrate that CMML significantly outperforms state-of-the-art (SOTA) methods, yielding AVG AUC improvements of 1.26\%, 0.97\%, and 1.32\%, respectively.
\end{abstract}




\begin{keywords}
Missing-modality learning\sep
Missing-modality completion\sep
Context token\sep
Memory bank\sep
Robust diagnosis
\end{keywords}

\maketitle

\section{Introduction}
\label{sec_introduction}
The clinical diagnosis of complex diseases relies on multimodal data comprising imaging and tabular records. However, obtaining complete sets is often challenging due to cost, patient conditions, or technical constraints. Recent studies~\citep{le2025multimodal, wang2024mgiml} demonstrate that models trained on complete data degrade significantly when certain modalities are absent. This vulnerability risks misdiagnosis, adversely affecting patient outcomes and prognosis. Therefore, developing a robust framework for accurate diagnosis under arbitrary missing conditions is essential.


Existing approaches to address this challenge follow two primary paradigms. The first paradigm~\citep{chen2019robust,wang2023multi,xu2024leveraging,li2025simmlm} focuses on extracting joint representations exclusively from the available modalities. Within this paradigm, some methods~\citep{chen2019robust,wang2023multi} employ feature disentanglement to capture modality-shared and modality-specific features, using shared information to compensate for missing data. Others~\citep{xu2024leveraging,li2025simmlm} explore inter-modal relationships to extract robust representations. Despite their progress, these methods inherently neglect the potential discriminative information latent in the missing modalities. By failing to account for the unique contributions of absent data, they inevitably yield suboptimal diagnostic performance, especially in complex clinical scenarios where multimodal synergy is critical.
The second paradigm emphasizes the completion of missing data at either the input or representation level. Input-level  completion~\citep{dorent2023unified,zhang2024unified} reconstructs raw pixels but often introduces excessive computational overhead and detrimental noise, which can impair downstream tasks. Consequently, representation-level  completion~\citep{guo2024multimodal, zhou2021latent} has gained traction due to its focus on high-level semantic information. Early efforts~\citep{guo2024multimodal, zhou2021latent} typically established specific mappings between modality pairs to recover missing latent features. More recently, researchers have shifted toward unified architectures\citep{sun2024redcore, huang2025mitigating} to capture inter-modal relationships within a single framework. Furthermore, synthesized representations often lack modality-specific details due to the inherent semantic gaps between heterogeneous modalities, and existing solutions~\citep{le2025multimodal, wang2024mgiml} are frequently limited to bi-modal settings, failing to generalize to more complex multimodal clinical scenarios.

By thoroughly analyzing current representation-level completion methods, we identify two critical challenges that remain unresolved. First, capturing intricate inter-modal dependencies under arbitrary missing scenarios is formidable; as the modality count increases, the combinatorial complexity of missing patterns grows exponentially, making it difficult to establish a stable inter-modal semantic consensus to guide the synthesis of absent representations without introducing semantic noise. Second, a profound hurdle arises during feature integration: the inherent distribution shift between original available and synthesized representations. Directly fusing these heterogeneous representations inevitably introduces a semantic modality gap, where synthesized representations lacking fine-grained modality-specific details can introduce semantic interference that impairs the discriminative integrity of the integrated multimodal representations.


In this paper, we propose an innovative Context-driven Missing-Modality Learning (CMML) framework aimed at achieving robust diagnosis under arbitrary missing conditions.
Specifically, we develop a Context-driven Modality Completion (CMC) module to address the first challenge. We first design a Cascade Residual Transformer-based Autoencoder (CRTA) equipped with learnable context tokens. These tokens act as dataset-level semantic prior to effectively capture inter-modal dependencies, enabling the CRTA to synthesize high-quality base representations for missing modalities. Subsequently, these synthesized representations are enriched with modality-specific details retrieved from dedicated memory banks.
To tackle the second challenge, we propose the Instance-adaptive Context-guided Alignment and Refinement (ICAR) module, which operates in a dual-stage manner. In the first stage, referred to as Instance-adaptive Context-guided Multimodal Alignment (ICMA), we adapt the context tokens initially capturing dataset-level semantic prior into instance-adaptive semantic references by infusing refined multimodal representations from the CRTA's outputs. Guided by each grounded reference, modality-specific cross-attention mechanisms project all representations into a unified space to resolve the semantic gap. In the second stage, the Class-aware Discriminative Refinement (CDR) strategy leverages contrastive learning to further extract diagnostic cues and enhance category separability. This integrated framework ensures that the final multimodal representations are both semantically consistent and highly optimized for accurate diagnosis.
The main contributions are summarized as follows:
\begin{itemize}
	\item We propose a novel CMML framework for robust disease diagnosis under arbitrary missing conditions. Specifically, CMC module leverages learnable context tokens and memory banks to synthesize essential representations for the missing modalities.
	\item We design ICAR module that establishes instance-adaptive semantic references to explicitly bridge the modality gap between original available and synthesized features.
	\item Extensive experiments on two public datasets and one in-house dataset demonstrate that the proposed framework significantly outperforms state-of-the-art (SOTA) methods in multimodal (image-tabular) diagnostic tasks across various arbitrary missing scenarios.
\end{itemize}

\section{Related Work}
Various approaches~\citep{wu2024deep} have been proposed to mitigate the performance degradation caused by missing modalities, which can be categorized into joint representation learning and multimodal information completion.

\begin{figure*}[ht]
    \centering
    \includegraphics[width=\linewidth]{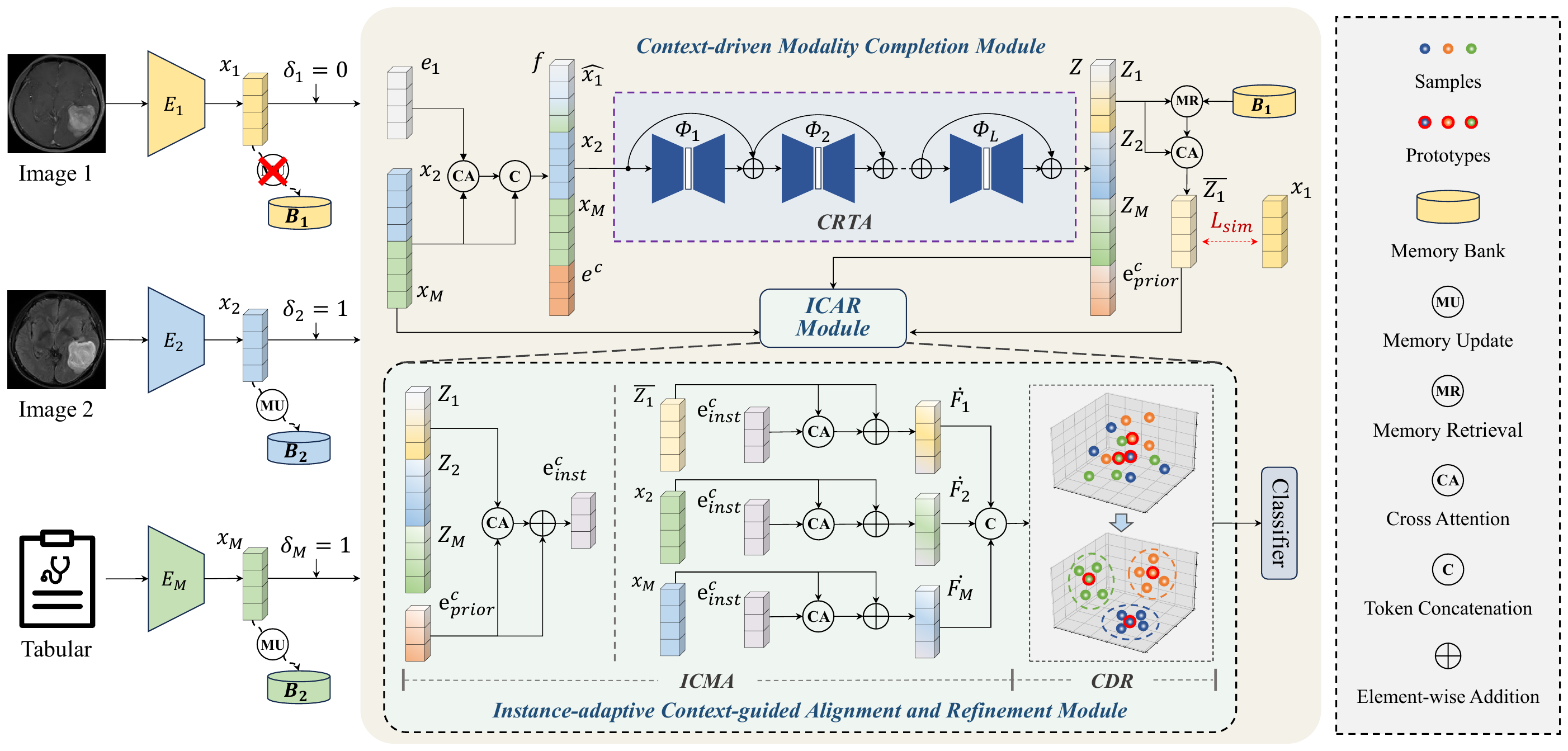}
    \caption{Overview of the proposed CMML framework. Initially, modality-specific features are extracted and processed by the Context-driven Modality Completion (CMC) module to synthesize representations for missing modalities. Subsequently, the Instance-adaptive Context-guided Alignment and Refinement (ICAR) module bridges the semantic gap between original available and synthesized features while extracting discriminative diagnostic information for the final prediction. 
    }
    \label{overview}
\end{figure*}
\subsection{Joint representation learning Approaches}
These methods focus on extracting robust multimodal features solely from available modalities. A primary strategy is feature disentanglement\citep{chen2019robust,wang2023multi}, which decomposes multimodal information into modality-shared and modality-specific components. For instance, ShaSpec\citep{wang2023multi} utilizes specialized loss functions to decouple multimodal information and use modality-shared feature to compensate for missing attributes.
Beyond feature-level decomposition, recent research emphasizes architectural robustness to partial inputs. Multimodal Transformers\citep{ma2022multimodal, zhang2022mmformer, liu2023sfusion, zhang2024tmformer} leverage self-attention mechanisms to capture long-range inter-modal dependencies, naturally tolerating missing data through attention masking or adaptive fusion strategies. Furthermore, Mixture of Experts (MoE) architectures\citep{xu2024leveraging, li2025simmlm} provide dynamic routing to handle modality fluctuations. Notably, SimMLM~\citep{li2025simmlm} introduces a Dynamic Mixture of Modality Experts (DMoME) to adaptively weigh the contribution of each modality and employs a More vs. Fewer ranking loss to maintain performance consistency across varying numbers of available inputs.
However, by inherently relying on restricted information and neglecting the discriminative cues latent in missing modalities, these methods inevitably yield suboptimal diagnostic performance~\citep{le2025multimodal, wang2024mgiml}.

\subsection{Multimodal Information Completion Approaches}
This paradigm focuses on synthesizing missing modality information and is generally bifurcated into input-level and representation-level completion.
Input-level completion reconstructs absent modalities in the pixel space by leveraging generative architectures such as GANs~\citep{zhou2020hi,zhang2024unified}, VAEs~\citep{wu2018multimodal,dorent2023unified}, and Diffusion Models~\citep{jiang2023cola,zhang2025foundation}.
Specifically, a GAN-based unified multimodal synthesis framework~\citep{zhang2024unified} leverages modality-invariant and specific features to handle arbitrary missingness, while MHVAE~\citep{dorent2023unified} employs a probabilistic formulation for joint image synthesis. More recently, the diffusion-based M2DN~\citep{meng2024multi} recovers missing inputs by treating them as random noise, and MoME~\citep{zhang2025foundation} utilizes a soft dispatch network to optimize the synthesis through weighted fusion of available modalities. However, these pixel-level reconstruction methods often incur heavy computational overhead and introduce artifacts that can conflict with the objectives of downstream diagnostic tasks.

To mitigate the impact of noise and focus on high-level semantic information, numerous studies have focused on generating latent representations for missing modalities~\citep{guo2024multimodal,tran2017missing,huang2025mitigating,wang2024mgiml}. Early efforts established deterministic mappings~\citep{guo2024multimodal} or learned statistical correlations~\citep{zhou2021latent} among modalities to recover missing representations. To better capture inter-modal relationships, cascaded residual autoencoders (CRA)~\citep{tran2017missing} and variational information bottleneck (VIB) methods~\citep{sun2024redcore} were introduced. Recently, Transformer-based frameworks have gained traction; for instance, high-order token interactions~\citep{huang2025mitigating} and cross-attention mechanisms~\citep{xu2026mcmoe} are employed to impute missing representations from available modality information. 
However, a significant modality gap often persists, causing synthesized features to lack modality-specific details. To address this, memory-driven strategies have been developed: MH-Complete~\citep{wang2024mgiml} utilizes modality-specific memory banks to reconstruct missing components, while disentanglement-based approaches~\citep{xiong2025disentanglement} employ codebooks to store shared and specific information. Despite their effectiveness, these methods are architecturally restricted to bi-modal scenarios. Their underlying designs, often predicated on pairwise interactions, lack the structural flexibility required to accommodate the arbitrary number of modalities typically encountered in complex clinical settings.
 
In contrast to these existing methods, our approach is capable of handling inputs with an arbitrary number of modalities. It generates high-quality representations for missing modalities by effectively capturing the relationships between modalities and utilizing memory bank which store modality-specific information for each modality.

\section{Method}
\label{section_method}
Let us denote a multimodal dataset comprising $N$ samples as $\{X^i,y^i\}_{i=1}^N$, where $y^i$ represents classification label of the $i$-th sample. Each sample consists of $M$ modalities, represented as $\{X^i_j\}_{j=1}^M$. Specifically, the first $M-1$ modalities are image types, represented as $\{X^i_m\}_{m=1}^{M-1}$, while the $M$-th modality corresponds to tabular data, denoted as $X^i_M$. 
It is assumed that all modalities are fully available during the training phase. For clarity of exposition, we describe the proposed framework using 2D images and tabular records as the representative inputs.

\subsection{Overview}
The proposed CMML framework is illustrated in Figure~\ref{overview}. Initially, multimodal tokens are extracted via modality-specific encoders. During training, a random modality dropout strategy is employed to enhance framework robustness. For available modalities, the extracted tokens are used to update modality-specific memory banks, which store representative information.
Missing representations are recovered through the CMC strategy. Specifically, learnable embeddings are concatenated with the available modality tokens and learnable context tokens to serve as input for the CRTA. Leveraging these context tokens as dataset-level semantic prior, the CRTA captures complex inter-modal dependencies via self-attention to synthesize base representations for the absent modalities, which are subsequently enriched with modality-specific details retrieved from the memory banks.
To address the modal gap between the original available and synthesized representations, we adapt the learned context tokens into instance-adaptive semantic references by infusing refined multimodal features harvested from the CRTA's outputs. Guided by each grounded reference, modality-specific cross-attention mechanisms project all representations into a unified space to achieve semantic alignment. Then, the CDR strategy leverages contrastive learning to extract diagnostic cues from the fused representations for final diagnostic task.

\subsection{Context-driven Modality Completion Module}
The CMC module recovers semantic representations of missing modalities within arbitrary frameworks. It comprises the CRTA architecture, facilitated by learnable context tokens that act as dataset-level semantic prior to capture inter-modal dependencies and modality-specific memory banks to provide latent information for representation enrichment.

\begin{table}[t]  
\centering  
\caption{Tabular attribute templates. Curly braces \{\} sequentially denote attribute names and values; numerical data are converted into English text.}  
\label{template}  
\resizebox{0.4\textwidth}{!}{  
\begin{tabular}{cc}  
\hline
\multirow{2}{*}{\textbf{Attributes}}    & \multirow{2}{*}{\textbf{Templates}}             \\
                                        &                                                 \\ \hline
management                              & The \{\} for the patient is \{\}                \\ \hdashline
sex                                     & \multirow{2}{*}{The \{\} of patient is \{\}}    \\
age                                     &                                                 \\ \hdashline
tumor area                              & \multirow{3}{*}{The \{\} in the brain is \{\}.} \\
edema area                              &                                                 \\
tumor location                          &                                                 \\ \hdashline
lesion location                         & \multirow{4}{*}{The \{\} is \{\}}               \\
lesion elevation                        &                                                 \\
level of diagnostic difficulty          &                                                 \\
value of apparent diffusion coefficient &                                                 \\ \hline
\end{tabular}}
\end{table}

\subsubsection{Feature Extraction}
For each image $X_m$, a specific encoder $E_m$ extracts feature map of shape $\mathbb{R}^{h\times w\times c}$ (where $c$, $h$ and $w$ denote channel dimension, height and width, respectively), which are embedded into tokens ($\mathbb{R}^{S_{m}\times c}$, where $S_m=h\times w$). Next, these tokens are projected to a unified dimension $d$ using a fully connected (FC) layer with ReLU activation.
For tabular data, attributes are first converted into textual descriptions via predefined templates (Table~\ref{template}), following~\citep{liu2026cfcml}.
To leverage its superior semantic understanding of clinical text, a frozen DeepSeek-R1 
model~\citep{guo2025deepseek} is then employed to extract tabular tokens with a shape of $\mathbb{R}^{S_M\times d_r}$, where $S_M$ and $d_r$ represents the token length and feature dimension. 
These tabular tokens are also projected to $d$ to ensure multimodal compatibility.
The token extraction from images and tabular data can be formulated as follows:
\begin{align}  
    x_m &= ReLU(FC(E_m(X_m))), \quad m \in \{1, \ldots, M-1\} \\
    x_M &= ReLU(FC(DeepSeek(X_M))).  
\end{align} 


\subsubsection{Multimodal Embedding with Arbitrary Missingness}
To ensure robustness under diverse modality absence scenarios, we employ a random modality dropout strategy during training. For each sample $i$, let $\delta_j\in\{0,1\}$ denote the availability of the $j$-th modality. We define $\mathcal{M}_{avail}=\{j\mid \delta_j=1\}$ and $\mathcal{M}_{miss}=\{j\mid \delta_j=0\}$ as the sets of available and missing modality indices, respectively. 

Typically, using a static learnable vector to represent missing data fails to account for inter-sample variability. To address this, we propose a dynamic initialization for the missing-modality embeddings. For an absent modality ($o \in \mathcal{M}_{miss}$), its initial representation $\hat{x}_o^i$ is synthesized by querying the available modalities:
\begin{equation}
  \hat{x}_o^i=CA(e_o, Concat(\{x_j^i \mid j\in \mathcal{M}_{avail}\})),
\end{equation}
where $Concat(\cdot)$ denotes the token concatenation operation, $CA(\cdot)$ denotes the cross-attention operation using a learnable modality seed $e_o$ as the Query and the concatenated available tokens as Key and Value. This design allows $\hat{x}_o^i$ to inherit a coarse semantic prior derived from the available modalities of the specific instance. 

To provide the CRTA with comprehensive structural and status information, we further enrich these representations with three types of modality-specific additive embeddings. Specifically, for each modality $j$, we define position embeddings 
$e_j^{pos}\in\mathbb{R}^{S_j\times d}$ to provide essential spatial or contextual cues, and modal-type embeddings~\citep{kim2021vilt} $e_j^{mt}\in\mathbb{R}^{S_j\times d}$ to distinguish between heterogeneous data sources (e.g., images and tabular data). Furthermore, we introduce missing-state embeddings $e_j^{ms}\in\mathbb{R}^{2\times S_j\times d}$ to explicitly signal the data availability status, where $e_j^{ms}[1]$ and $e_j^{ms}[0]$ denote the available and missing states, respectively. 
By incorporating these informative cues, the final input representation $f_j$ is formulated as:
\begin{equation}
   f_j = \tilde{x}_j + e^{pos} + e^{mt} + e^{ms}[\delta_j], \quad
   \tilde{x}_j =
   \begin{cases}
   x_j, & \delta_j = 1 \\
   \hat{x}_j, & \delta_j = 0
   \end{cases}
\end{equation}

\begin{figure}
    \centering
    \includegraphics[width=\linewidth]{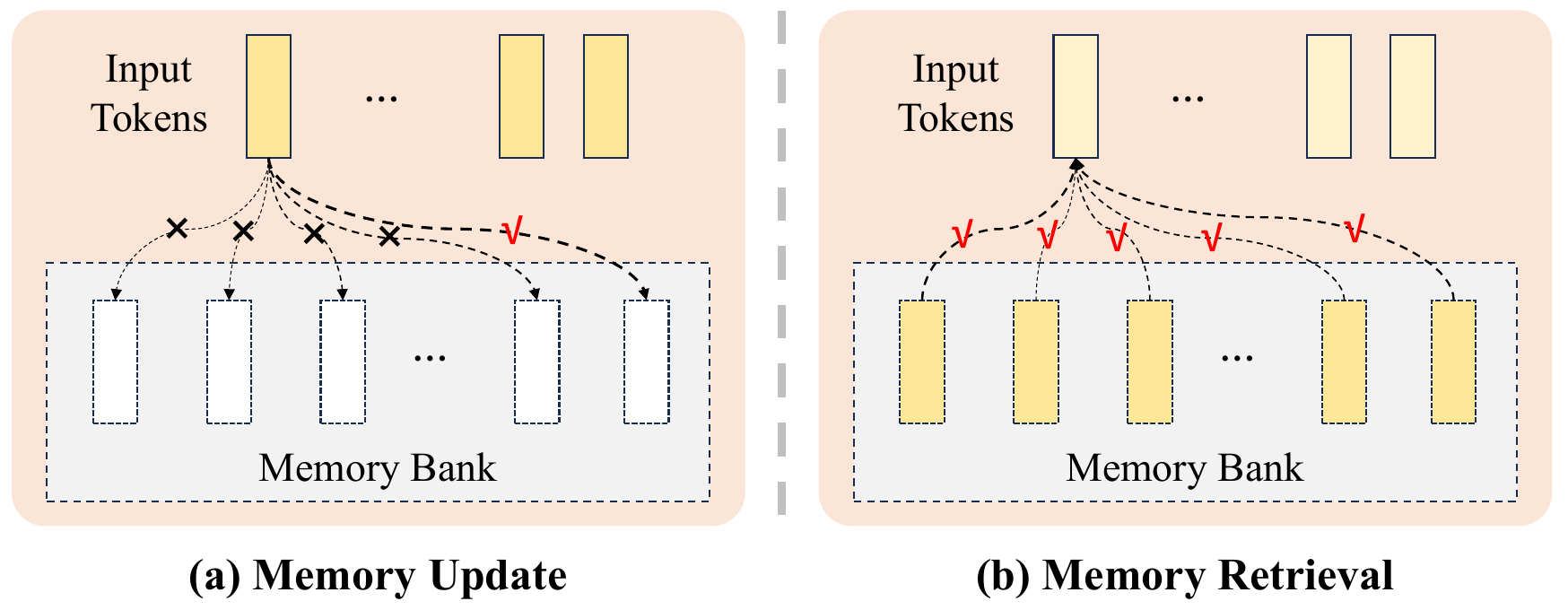}
    \caption{The process of the memory update and retrieval. In the update phase, each input token updates the memory token with the highest similarity. In the retrieval phase, all memory tokens enhance the input token based on similarity weights.}
    \label{memory_op}
\end{figure}

\subsubsection{Context-driven Missing Representation Synthesis}
Inspired by CRA~\citep{tran2017missing}, we propose the CRTA to synthesize missing representations by capturing high-order inter-modal dependencies. As illustrated in Figure~\ref{overview}, the CRTA consists of $L$ RTAs, denoted as $\Phi_l$ where $l\in\{1, .., L\}$  (where $L$ is set to 5), to progressively recover the latent representations of missing modalities. 
Each RTA follows an encoder-decoder bottleneck design, with both components built upon the standard Transformer blocks. 
Specifically, the dimensionality transformation is performed within the Transformer’s internal Query-Key-Value projection layers, compressing the input to $d/r$ (where $r = 4$) in the encoder and restoring it to $d$ in the decoder.
By leveraging these bottlenecks and residual learning, the CRTA decomposes the complex synthesis task into incremental stages, facilitating the robust recovery of essential semantic information for the absent modalities.

Inspired by the extra token introduced in~\citep{caron2021emerging}, we present the learnable context tokens $e^{c}\in \mathbb{R}^{N_{ct}\times d}$ to capture inter-modal dependencies, where $N_{ct}$ denotesdenotes the number of context tokens. 
These tokens are concatenated with multimodal tokens $f_j$ as the input $f = Concat(e^c, f_1, ..., f_M)$ for the CRTA. The cascaded synthesis process is formulated as:

\begin{equation}
\begin{cases}
   z_l = \Phi_l(f), & l = 1 \\
   z_l = \Phi_l\left(f + \sum_{k=1}^{l-1} z_k\right), & l > 1
\end{cases}  
\end{equation}
where $z_l$ denotes the output of $l$-th TRA. We denote $Z$ as the final output of the CRTA. Specifically, the updated context tokens $e^c_{prior}=Z_0$ and unimodal tokens $Z_j$ are extracted from $Z$ at positions consistent with their original arrangement in the input sequence $f$.

Crucially, the context tokens function as dataset-level semantic prior during the self-attention process within the CRTA. By attending to all available modality tokens, they aggregate a multimodal consensus that encapsulates the latent inter-modal dependencies across the entire training distribution. This learned prior provides strong semantic constraints and guidance for the synthesis of absent modalities, ensuring that the recovered representations conform to the global statistical patterns established by the dataset.

\subsubsection{Memory-driven Representation Enrichment}
While the CRTA reconstructs base semantic features, the inherent modality gap often leads to a loss of fine-grained, modality-specific details. To address this, we establish a modality-specific memory bank~\citep{huang2021memory} $B_j = \{B_{j,b}\}_{b=1}^{N_t}$ for each modality to store representative information, where $N_t$ denotes the token number in the memory bank. 

\textbf{Memory Update}: During memory update epoch $N_e$ in training process, when the $j$-th modality is available, its extracted tokens $x_j$ are used to update $B_j$ via a moving average strategy. For each extracted token $x_{j,k}$, we identify the most similar memory token $B_{j,b^*}$ and update it:
\begin{equation}
    \begin{cases}
        b^* = \mathop{\arg\max}\limits_{b} CS(x_{j,k}, B_{j,b}), \\
        B_{j,b^*} \leftarrow (1-\lambda) B_{j,b^*} + \lambda x_{j,k},
    \end{cases}
\end{equation}
where $CS(\cdot)$ denotes cosine similarity operation, and $\lambda\in[0,1]$ is the decay rate, which is experimentally set to $0.2$.

\textbf{Memory Retrieval}: For an absent modality $o$, each base token $Z_{o,k}$ is enriched based on the corresponding memory bank $B_o$ via a soft-attention retrieval strategy. The retrieval of modality-specific information $G_o$ and the construction of the final completed representation $\bar{Z}_o$ are formulated as:
\begin{equation}
    \begin{cases}
    w_o^{kb} = Softmax\left( \left[ CS(Z_{o,k}, B_{o,b}) \right]_{b=1}^{N_t} \right), \\
    G_o^k = \sum_{b=1}^{N_t} w_o^{kb} B_{o,b}, \\
    \bar{Z}_o = Z_o + MHCA(Z_o, G_o),
    \end{cases}
\end{equation}
where $MHCA(,)$ denotes the multi-head cross-attention (MHCA) layer that adaptively infuses the retrieved modality-specific information into the synthesized base features to recover fine-grained details.

To supervise the synthesis process, we employ a similarity loss 
$\mathcal{L}_{sim}$ during training, encouraging the synthesized features $\bar{Z}_o$ to align with the original features $x_o$:
\begin{equation}
    \mathcal{L}_{sim} = 1 - \frac{1}{S_o} \sum_{k=1}^{S_o} CS(x_{o,k}, \bar{Z}_{o,k}).
\end{equation}

\subsection{Instance-adaptive Context-guided Alignment and Refinement Module}
To resolve the semantic gap between original available and synthesized representations, the ICAR module sequentially performs multimodal alignment (ICMA) and discriminative refinement (CDR) as follows.

\subsubsection{Instance-adaptive Context-guided Multimodal Alignment Module}
The ICMA module is designed to achieve multimodal alignment through a two-step procedure: initially formulating instance-adaptive semantic references and subsequently projecting all representations from each instance toward the corresponding reference.

\textbf{Reference Formulation:} 
In this step, we adapt the context tokens with refined representations from the CRTA'outputs. Since these refined representations have undergone high-order inter-modal interactions in CRTA, they encapsulate a multimodal consensus that uniquely represents the current instance’s semantic profile. This process effectively grounds the dataset-level prior with each instance-specific evidence, yielding an instance-adaptive semantic reference $e^c_{inst}$ formulated as:
\begin{equation}
    e^c_{inst} = e^c_{prior}+CA(e^c_{prior}, Cat(Z_1,...,Z_M)).
\end{equation}

\textbf{Representation Alignment:}
We then utilize $e^c_{inst}$ as a shared reference to bridge the modality gap. For the completed feature set $\mathcal{F}=\{x_1,...,x_{o-1},\bar{Z}_o,$ $x_{o+1},..., x_M\}$ where the $o$-th modality is absent, each representation $\mathcal{F}_j$ is aligned toward this reference via a cross-attention mechanism, yielding each harmonized representation $F_j$:
\begin{equation}
    F_j = \mathcal{F}_j + MHCA(\mathcal{F}_j, e^c_{inst}).
\end{equation}
By projecting disparate modalities into this unified semantic space, the ICMA module effectively resolves distribution discrepancies and ensures semantic consistency.

\subsubsection{Class-aware Discriminative Refinement Strategy}
Building upon the aligned representations, the CDR strategy is introduced to further extract diagnostic cues and enhance category separability. We first integrate the unimodal features via concatenation and global average pooling (GAP) to form the fused multimodal feature $\dot{F}$. The prediction $y'$ is generated through an MLP-based classifier and optimized by the cross-entropy (CE) loss:
\begin{equation}
\begin{cases}
    \dot{F} = GAP(Concat(F_1,...,F_M)),\\
    y' = MLP(\dot{F}),\\
    \mathcal{L}_{ce} = CE(y',y).
\end{cases}
\end{equation}

To foster intra-class compactness, we define a class prototype $P^{cls}$ for each category $cls$ within a batch of $N_b$ samples:
\begin{equation} 
    P^{cls} = \frac{\sum_{i=1}^{N_b} \dot{F}^i \mathbbm{1}[y^{i} = cls]}{\sum_{i=1}^{N_b} \mathbbm{1}[y^{i}=cls]}.
\end{equation}
We then implement dual-level contrastive learning (CL) using a generalized loss $\mathcal{L}_{cl}$ formulated over an anchor set $\mathcal{A}=\{A_i\}_{i=1}^{N_A}$, where $N_A$ represents the capacity of the anchor set. For each anchor $A_i$, let $\mathcal{S}_i^+$ and $\mathcal{S}_i^-$ denote its corresponding positive and negative sets:
\begin{equation}
    \mathcal{L}_{cl} = -\frac{1}{N_A} \sum_{i=1}^{N_A} \log \frac{\sum_{s \in \mathcal{S}^+_i} exp(CS(A_i, s) / \tau)}{\sum_{s \in \mathcal{S}_i} exp(CS(A_i, s) / \tau)},
    \label{Eq_cl}
\end{equation}
where $\mathcal{S}_i=\mathcal{S}^+_i\cup \mathcal{S}^-_i$, $\tau$ is the temperature parameter. Specifically, we derive two contrastive losses, $\mathcal{L}^{sam}_{cl}$ and $\mathcal{L}^{proto}_{cl}$, by varying the anchor and set compositions in Eq.~\ref{Eq_cl}:
\newline\textbf{1) Sample-based CL}($\mathcal{L}^{sam}_{cl}$): Each sample serves as an anchor, while its positive set contains the corresponding prototype and negative set contains the other prototypes.
\newline\textbf{2) Prototype-based CL}($\mathcal{L}^{proto}_{cl}$): Each class prototype serves as an anchor, while its positive set contains samples sharing the same label, and negative set contains remaining prototypes.

Finally, the overall loss $\mathcal{L}$ is characterized as:
\begin{equation}
\mathcal{L} = \mathcal{L}_{ce} + \alpha\mathcal{L}_{sim}+ \beta\mathcal{L}^{sam}_{cl}+\gamma\mathcal{L}^{proto}_{cl},
\end{equation}
where $\alpha$, $\beta$ and $\gamma$ are balancing hyperparameters.
\begin{figure}
    \centering    
    \includegraphics[width=\linewidth]{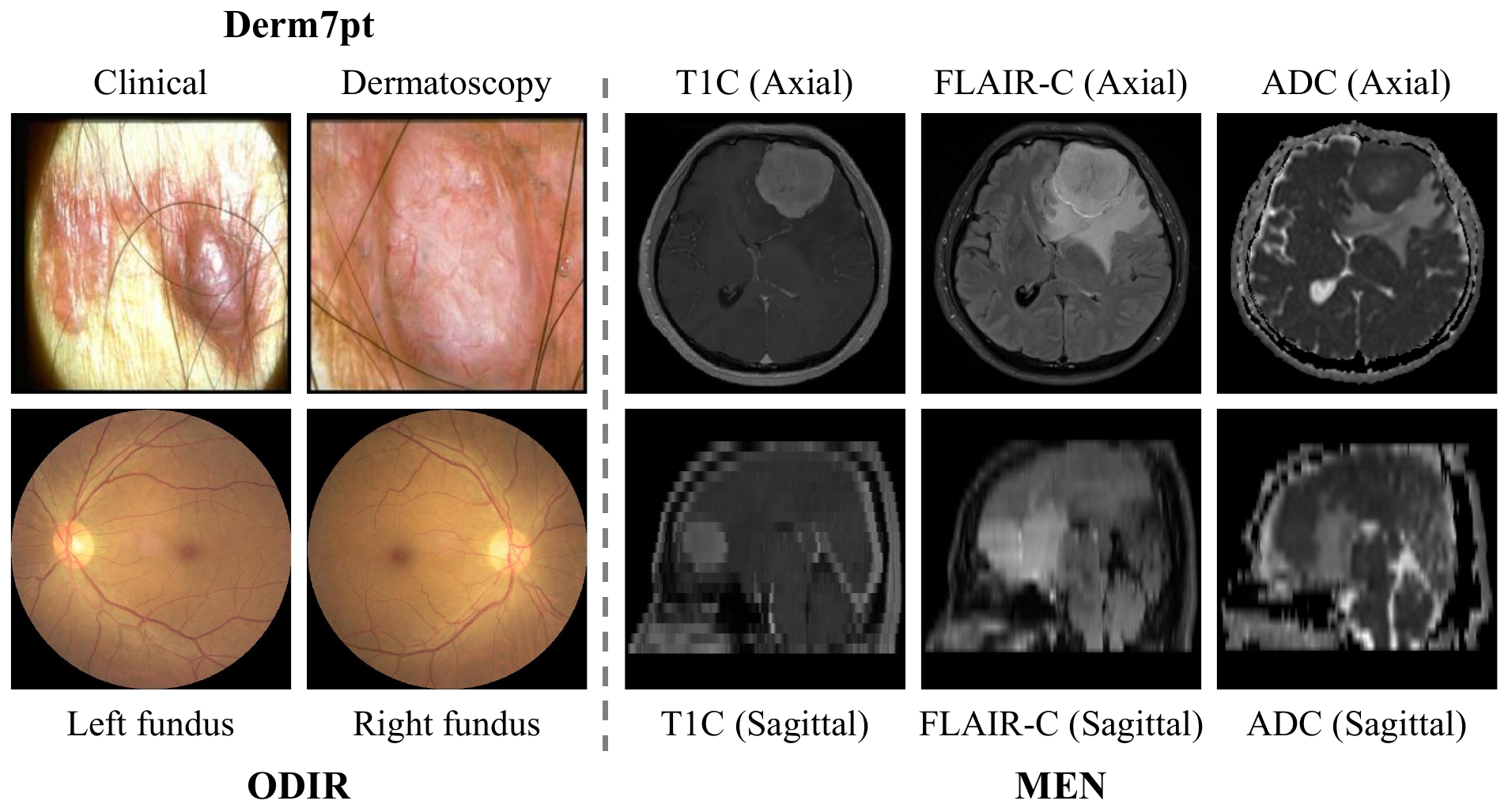}
    \caption{Representative multimodal samples from the three evaluated datasets. For the MEN dataset, 3D MRI sequences are illustrated from orthogonal axial and sagittal planes to display the full spatial extent of the volumetric data.}
    \label{data_show}
\end{figure}
\begin{table*}
\centering
\caption{The comparison results on the Derm7pt and ODIR datasets (mean$\pm$standard deviation, values are percentages, \%). The best results are highlighted in \textcolor{red}{red}, while the second-best results are indicted in \textbf{bold}. Statistical significance (p-value<0.05) in comparison to other methods is indicated by *.The notation in the "Available Modalities" is defined as follows: for Derm7pt, 0: clinical image, 1: dermatoscopic image, 2: tabular data; for ODIR, 0: left fundus image, 1: right fundus image, 2: tabular data. 'AVG' represents the average result across all modality conditions.}
\label{result_comparison}
\resizebox{\textwidth}{!}{
\Large
\begin{tabular}{c|c|cccccccccccccc|cc} 
\hline
\multirow{3}{*}{Datasets} & \multirow{3}{*}{Methods} & \multicolumn{14}{c|}{Available Modalities}                                                                                                                                          & \multicolumn{2}{c}{\multirow{2}{*}{AVG}}  \\ 
\cline{3-16}
                          &                          & \multicolumn{2}{c}{0}   & \multicolumn{2}{c}{1}   & \multicolumn{2}{c}{2}   & \multicolumn{2}{c}{01}   & \multicolumn{2}{c}{02}   & \multicolumn{2}{c}{12}  & \multicolumn{2}{c|}{012} & \multicolumn{2}{c}{}                      \\ 
\cline{3-18}
                          &                          & ACC        & AUC        & ACC        & AUC        & ACC        & AUC        & ACC        & AUC        & ACC        & AUC        & ACC        & AUC        & ACC        & AUC        & ACC         & AUC                         \\ 
\hline
\multirow{10}{*}{Derm7pt}  & ShaSpec~\cite{wang2023multi}                  & \makecell[c]{\vspace{-1mm}71.25* \\ \large $ \pm $2.69}                           & \makecell[c]{\vspace{-1mm}72.94* \\ \large $ \pm $3.50}                           & \makecell[c]{\vspace{-1mm}71.46* \\ \large $ \pm $2.30}                           & \makecell[c]{\vspace{-1mm}80.46* \\ \large $ \pm $0.99}                             & \makecell[c]{\vspace{-1mm}72.08* \\ \large $ \pm $3.37}                                    & \makecell[c]{\vspace{-1mm}\textbf{\textcolor{red}{89.05}} \\ \large $ \pm $0.43}  & \makecell[c]{\vspace{-1mm}77.40* \\ \large $ \pm $1.45}                           & \makecell[c]{\vspace{-1mm}83.23* \\ \large $ \pm $2.22}                           & \makecell[c]{\vspace{-1mm}75.73* \\ \large $ \pm $3.43}                           & \makecell[c]{\vspace{-1mm}\textbf{\textcolor{red}{89.47}} \\ \large $ \pm $1.33} & \makecell[c]{\vspace{-1mm}79.90* \\ \large $ \pm $1.62}                           & \makecell[c]{\vspace{-1mm}90.99 \\ \large $ \pm $0.72}                           & \makecell[c]{\vspace{-1mm}81.88* \\ \large $ \pm $1.11}                           & \makecell[c]{\vspace{-1mm}\textbf{91.12}\\ \large $ \pm $1.34}                  & \makecell[c]{\vspace{-1mm}75.67*\\ \large $ \pm $3.96}                            & \makecell[c]{\vspace{-1mm}85.32\\ \large $ \pm $6.30}                            \\
                          & SFusion~\cite{liu2023sfusion}                 & \makecell[c]{\vspace{-1mm}74.06 \\ \large $ \pm $0.26}                           & \makecell[c]{\vspace{-1mm}78.43* \\ \large $ \pm $2.42}                           & \makecell[c]{\vspace{-1mm}\textbf{81.14}* \\ \large $ \pm $0.82}                  & \makecell[c]{\vspace{-1mm}84.94* \\ \large $ \pm $0.62}                             & \makecell[c]{\vspace{-1mm}71.15* \\ \large $ \pm $0.39}                                    & \makecell[c]{\vspace{-1mm}82.22* \\ \large $ \pm $1.00}                            & \makecell[c]{\vspace{-1mm}\textbf{\textcolor{red}{82.71}} \\ \large $ \pm $0.97} & \makecell[c]{\vspace{-1mm}86.43* \\ \large $ \pm $0.64}                           & \makecell[c]{\vspace{-1mm}75.63* \\ \large $ \pm $2.43}                           & \makecell[c]{\vspace{-1mm}86.64 \\ \large $ \pm $2.10}                           & \makecell[c]{\vspace{-1mm}\textbf{83.54}* \\ \large $ \pm $0.39}                  & \makecell[c]{\vspace{-1mm}\textbf{91.00}* \\ \large $ \pm $0.28}                  & \makecell[c]{\vspace{-1mm}\textbf{84.06} \\ \large $ \pm $0.51}                  & \makecell[c]{\vspace{-1mm}91.04* \\ \large $ \pm $0.38}                           & \makecell[c]{\vspace{-1mm}\textbf{78.90}*\\ \large $ \pm $4.81}                   & \makecell[c]{\vspace{-1mm}85.81* \\ \large $ \pm $4.20}                             \\
                          & Redcore~\cite{sun2024redcore}                 & \makecell[c]{\vspace{-1mm}63.02* \\ \large $ \pm $3.84}                           & \makecell[c]{\vspace{-1mm}56.18* \\ \large $ \pm $7.51}                           & \makecell[c]{\vspace{-1mm}76.88* \\ \large $ \pm $3.60}                           & \makecell[c]{\vspace{-1mm}84.54* \\ \large $ \pm $0.70}                             & \makecell[c]{\vspace{-1mm}68.54* \\ \large $ \pm $2.30}                                    & \makecell[c]{\vspace{-1mm}79.94* \\ \large $ \pm $1.07}                            & \makecell[c]{\vspace{-1mm}76.77* \\ \large $ \pm $3.76}                           & \makecell[c]{\vspace{-1mm}82.87* \\ \large $ \pm $3.10}                           & \makecell[c]{\vspace{-1mm}68.85* \\ \large $ \pm $2.37}                           & \makecell[c]{\vspace{-1mm}80.30* \\ \large $ \pm $1.46}                           & \makecell[c]{\vspace{-1mm}74.06* \\ \large $ \pm $2.43}                           & \makecell[c]{\vspace{-1mm}89.14* \\ \large $ \pm $0.65}                           & \makecell[c]{\vspace{-1mm}76.04* \\ \large $ \pm $2.08}                           & \makecell[c]{\vspace{-1mm}88.81* \\ \large $ \pm $0.58}                           & \makecell[c]{\vspace{-1mm}72.02* \\ \large $ \pm $4.92}                            & \makecell[c]{\vspace{-1mm}80.25* \\ \large $ \pm $10.40}                            \\
                          & SimMLM~\cite{li2025simmlm}                    & \makecell[c]{\vspace{-1mm}\textbf{\textcolor{red}{77.50}} \\ \large $ \pm $0.51} & \makecell[c]{\vspace{-1mm}\textbf{80.21}* \\ \large $ \pm $0.83}                  & \makecell[c]{\vspace{-1mm}72.29* \\ \large $ \pm $6.16}                           & \makecell[c]{\vspace{-1mm}\textbf{86.83}* \\ \large $ \pm $0.53}                    & \makecell[c]{\vspace{-1mm}64.06* \\ \large $ \pm $0.01}                                    & \makecell[c]{\vspace{-1mm}80.72* \\ \large $ \pm $0.13}                            & \makecell[c]{\vspace{-1mm}78.13* \\ \large $ \pm $4.21}                           & \makecell[c]{\vspace{-1mm}\textbf{86.56}* \\ \large $ \pm $0.88}                  & \makecell[c]{\vspace{-1mm}77.61* \\ \large $ \pm $0.39}                           & \makecell[c]{\vspace{-1mm}80.53* \\ \large $ \pm $0.91}                           & \makecell[c]{\vspace{-1mm}72.71* \\ \large $ \pm $5.81}                           & \makecell[c]{\vspace{-1mm}87.04* \\ \large $ \pm $0.46}                           & \makecell[c]{\vspace{-1mm}79.79* \\ \large $ \pm $2.75}                           & \makecell[c]{\vspace{-1mm}86.91* \\ \large $ \pm $0.88}                           & \makecell[c]{\vspace{-1mm}74.58* \\ \large $ \pm $5.03}                            & \makecell[c]{\vspace{-1mm}84.11* \\ \large $ \pm $3.15}                             \\
                          & MCMoE~\cite{xu2026mcmoe}                      & \makecell[c]{\vspace{-1mm}72.50* \\ \large $ \pm $2.46}                           & \makecell[c]{\vspace{-1mm}76.64* \\ \large $ \pm $1.96}                           & \makecell[c]{\vspace{-1mm}77.50* \\ \large $ \pm $0.77}                           & \makecell[c]{\vspace{-1mm}83.58* \\ \large $ \pm $0.84}                             & \makecell[c]{\vspace{-1mm}\textbf{\textcolor{red}{76.35}} \\ \large $ \pm $7.47}          & \makecell[c]{\vspace{-1mm}\textbf{87.50} \\ \large $ \pm $1.24}                   & \makecell[c]{\vspace{-1mm}78.85* \\ \large $ \pm $1.70}                           & \makecell[c]{\vspace{-1mm}85.23* \\ \large $ \pm $1.46}                           & \makecell[c]{\vspace{-1mm}\textbf{78.34} \\ \large $ \pm $2.06}                  & \makecell[c]{\vspace{-1mm}\textbf{88.83} \\ \large $ \pm $0.83}                  & \makecell[c]{\vspace{-1mm}81.87* \\ \large $ \pm $0.76}                           & \makecell[c]{\vspace{-1mm}90.95* \\ \large $ \pm $0.46}                           & \makecell[c]{\vspace{-1mm}82.50* \\ \large $ \pm $1.11}                           & \makecell[c]{\vspace{-1mm}90.89* \\ \large $ \pm $0.35}                           & \makecell[c]{\vspace{-1mm}78.27* \\ \large $ \pm $3.39}                            & \makecell[c]{\vspace{-1mm}\textbf{86.23}\\ \large $ \pm $5.04}                    \\
\cline{2-18}
                          & Ours                                          & \makecell[c]{\vspace{-1mm}\textbf{75.73} \\ \large $ \pm $3.70}                  & \makecell[c]{\vspace{-1mm}\textbf{\textcolor{red}{81.68}} \\ \large $ \pm $1.50} & \makecell[c]{\vspace{-1mm}\textbf{\textcolor{red}{82.81}} \\ \large $ \pm $1.77} & \makecell[c]{\vspace{-1mm}\textbf{\textcolor{red}{87.95}} \\ \large $ \pm $0.50}   & \makecell[c]{\vspace{-1mm}\textbf{73.75} \\ \large $ \pm $0.89}                           & \makecell[c]{\vspace{-1mm}83.17 \\ \large $ \pm $1.95}                            & \makecell[c]{\vspace{-1mm}\textbf{82.50} \\ \large $ \pm $1.17}                  & \makecell[c]{\vspace{-1mm}\textbf{\textcolor{red}{89.42}} \\ \large $ \pm $0.79} & \makecell[c]{\vspace{-1mm}\textbf{\textcolor{red}{79.90}} \\ \large $ \pm $1.98} & \makecell[c]{\vspace{-1mm}87.41 \\ \large $ \pm $0.84}                           & \makecell[c]{\vspace{-1mm}\textbf{\textcolor{red}{84.58}} \\ \large $ \pm $1.21} & \makecell[c]{\vspace{-1mm}\textbf{\textcolor{red}{91.22}} \\ \large $ \pm $0.45} & \makecell[c]{\vspace{-1mm}\textbf{\textcolor{red}{84.69}} \\ \large $ \pm $0.92} & \makecell[c]{\vspace{-1mm}\textbf{\textcolor{red}{91.61}}\\ \large $ \pm $0.27} & \makecell[c]{\vspace{-1mm}\textbf{\textcolor{red}{80.57}}\\ \large $ \pm $4.00}  &  \makecell[c]{\vspace{-1mm}\textbf{\textcolor{red}{87.49}}\\ \large $ \pm $3.53}  \\
\hline
\multirow{10}{*}{ODIR}     & ShaSpec~\cite{wang2023multi}                   & \makecell[c]{\vspace{-1mm}59.10* \\ \large $ \pm $5.29}                           & \makecell[c]{\vspace{-1mm}80.70* \\ \large $ \pm $0.81}                           & \makecell[c]{\vspace{-1mm}48.16* \\ \large $ \pm $4.89}                           & \makecell[c]{\vspace{-1mm}81.26* \\ \large $ \pm $0.14}                           & \makecell[c]{\vspace{-1mm}33.66* \\ \large $ \pm $6.59}                           & \makecell[c]{\vspace{-1mm}51.98* \\ \large $ \pm $1.14}                           & \makecell[c]{\vspace{-1mm}58.35* \\ \large $ \pm $7.34}                           & \makecell[c]{\vspace{-1mm}87.16* \\ \large $ \pm $0.61}                           & \makecell[c]{\vspace{-1mm}60.92* \\ \large $ \pm $2.90}                           & \makecell[c]{\vspace{-1mm}79.78* \\ \large $ \pm $1.21}                           & \makecell[c]{\vspace{-1mm}56.60* \\ \large $ \pm $9.18}                           & \makecell[c]{\vspace{-1mm}81.01* \\ \large $ \pm $0.35}                           & \makecell[c]{\vspace{-1mm}64.75* \\ \large $ \pm $6.88}                           & \makecell[c]{\vspace{-1mm}87.20* \\ \large $ \pm $0.61}                           & \makecell[c]{\vspace{-1mm}54.51* \\ \large $ \pm $9.72}                            & \makecell[c]{\vspace{-1mm}78.44* \\ \large $ \pm $11.18}                                       \\
                          & SFusion~\cite{liu2023sfusion}                  & \makecell[c]{\vspace{-1mm}69.86* \\ \large $ \pm $1.05}                           & \makecell[c]{\vspace{-1mm}\textbf{84.38}* \\ \large $ \pm $0.68}                  & \makecell[c]{\vspace{-1mm}69.71* \\ \large $ \pm $2.09}                           & \makecell[c]{\vspace{-1mm}\textbf{\textcolor{red}{85.30}} \\ \large $ \pm $1.07} & \makecell[c]{\vspace{-1mm}36.23* \\ \large $ \pm $2.15}                           & \makecell[c]{\vspace{-1mm}53.16* \\ \large $ \pm $0.22}                           & \makecell[c]{\vspace{-1mm}\textbf{79.63}* \\ \large $ \pm $1.07}                  & \makecell[c]{\vspace{-1mm}\textbf{92.06} \\ \large $ \pm $0.62}                  & \makecell[c]{\vspace{-1mm}\textbf{70.28}* \\ \large $ \pm $0.68}                  & \makecell[c]{\vspace{-1mm}\textbf{84.48} \\ \large $ \pm $0.69}                  & \makecell[c]{\vspace{-1mm}70.81* \\ \large $ \pm $0.91}                           & \makecell[c]{\vspace{-1mm}\textbf{\textcolor{red}{85.27}} \\ \large $ \pm $0.97} & \makecell[c]{\vspace{-1mm}\textbf{79.97}* \\ \large $ \pm $0.71}                  & \makecell[c]{\vspace{-1mm}\textbf{92.12}\\ \large $ \pm $0.59}                  & \makecell[c]{\vspace{-1mm}\textbf{68.07}*\\ \large $ \pm $13.67}                  & \makecell[c]{\vspace{-1mm}\textbf{82.40}\\ \large $ \pm $12.35}                               \\
                          & Redcore~\cite{sun2024redcore}                  & \makecell[c]{\vspace{-1mm}66.60* \\ \large $ \pm $0.49}                           & \makecell[c]{\vspace{-1mm}81.90* \\ \large $ \pm $0.72}                           & \makecell[c]{\vspace{-1mm}69.29* \\ \large $ \pm $0.80}                           & \makecell[c]{\vspace{-1mm}81.81* \\ \large $ \pm $0.97}                           & \makecell[c]{\vspace{-1mm}\textbf{37.41}* \\ \large $ \pm $2.15}                  & \makecell[c]{\vspace{-1mm}50.14* \\ \large $ \pm $1.57}                           & \makecell[c]{\vspace{-1mm}76.22* \\ \large $ \pm $1.13}                           & \makecell[c]{\vspace{-1mm}88.62* \\ \large $ \pm $0.70}                           & \makecell[c]{\vspace{-1mm}66.19* \\ \large $ \pm $1.59}                           & \makecell[c]{\vspace{-1mm}80.79* \\ \large $ \pm $0.54}                           & \makecell[c]{\vspace{-1mm}69.82* \\ \large $ \pm $1.04}                           & \makecell[c]{\vspace{-1mm}82.19* \\ \large $ \pm $0.89}                           & \makecell[c]{\vspace{-1mm}75.92* \\ \large $ \pm $0.59}                           & \makecell[c]{\vspace{-1mm}88.32* \\ \large $ \pm $0.54}                           & \makecell[c]{\vspace{-1mm}65.92* \\ \large $ \pm $12.22}                           & \makecell[c]{\vspace{-1mm}79.11* \\ \large $ \pm $12.20}                                        \\
                          & SimMLM~\cite{li2025simmlm}                     & \makecell[c]{\vspace{-1mm}\textbf{70.05}* \\ \large $ \pm $1.09}                  & \makecell[c]{\vspace{-1mm}84.33 \\ \large $ \pm $0.86}                           & \makecell[c]{\vspace{-1mm}\textbf{71.00}* \\ \large $ \pm $0.57}                  & \makecell[c]{\vspace{-1mm}84.50* \\ \large $ \pm $0.67}                           & \makecell[c]{\vspace{-1mm}33.43* \\ \large $ \pm $1.20}                           & \makecell[c]{\vspace{-1mm}\textbf{\textcolor{red}{54.90}} \\ \large $ \pm $0.94} & \makecell[c]{\vspace{-1mm}77.62* \\ \large $ \pm $0.61}                           & \makecell[c]{\vspace{-1mm}90.64* \\ \large $ \pm $0.59}                           & \makecell[c]{\vspace{-1mm}70.05* \\ \large $ \pm $1.09}                           & \makecell[c]{\vspace{-1mm}84.30 \\ \large $ \pm $0.83}                           & \makecell[c]{\vspace{-1mm}\textbf{71.00}* \\ \large $ \pm $0.57}                  & \makecell[c]{\vspace{-1mm}84.58 \\ \large $ \pm $0.68}                           & \makecell[c]{\vspace{-1mm}78.00* \\ \large $ \pm $0.84}                           & \makecell[c]{\vspace{-1mm}90.70* \\ \large $ \pm $0.46}                           & \makecell[c]{\vspace{-1mm}67.31* \\ \large $ \pm $14.20}                           & \makecell[c]{\vspace{-1mm}81.99\\ \large $ \pm $11.39}                                        \\
                          & MCMoE~\cite{xu2026mcmoe}                       & \makecell[c]{\vspace{-1mm}69.33* \\ \large $ \pm $1.14}                           & \makecell[c]{\vspace{-1mm}84.00 \\ \large $ \pm $0.20}                           & \makecell[c]{\vspace{-1mm}69.97* \\ \large $ \pm $2.98}                           & \makecell[c]{\vspace{-1mm}84.62 \\ \large $ \pm $1.15}                           & \makecell[c]{\vspace{-1mm}42.10* \\ \large $ \pm $0.04}                           & \makecell[c]{\vspace{-1mm}51.65* \\ \large $ \pm $0.68}                           & \makecell[c]{\vspace{-1mm}77.58* \\ \large $ \pm $1.51}                           & \makecell[c]{\vspace{-1mm}91.02* \\ \large $ \pm $0.24}                           & \makecell[c]{\vspace{-1mm}68.88* \\ \large $ \pm $1.48}                           & \makecell[c]{\vspace{-1mm}84.06* \\ \large $ \pm $0.20}                           & \makecell[c]{\vspace{-1mm}68.99* \\ \large $ \pm $3.54}                           & \makecell[c]{\vspace{-1mm}84.53 \\ \large $ \pm $1.18}                           & \makecell[c]{\vspace{-1mm}77.40* \\ \large $ \pm $0.99}                           & \makecell[c]{\vspace{-1mm}90.90* \\ \large $ \pm $0.25}                          & \makecell[c]{\vspace{-1mm}67.77* \\ \large $ \pm $11.96}                            & \makecell[c]{\vspace{-1mm}81.54* \\ \large $ \pm $13.55}                                        \\
\cline{2-18}
                          & Ours                     & \makecell[c]{\vspace{-1mm}\textbf{\textcolor{red}{71.34}} \\ \large $ \pm $1.13} & \makecell[c]{\vspace{-1mm}\textbf{\textcolor{red}{84.59}} \\ \large $ \pm $1.27} & \makecell[c]{\vspace{-1mm}\textbf{\textcolor{red}{72.40}} \\ \large $ \pm $1.28} & \makecell[c]{\vspace{-1mm}\textbf{85.07} \\ \large $ \pm $1.16}                  & \makecell[c]{\vspace{-1mm}\textbf{\textcolor{red}{43.16}} \\ \large $ \pm $0.20} & \makecell[c]{\vspace{-1mm}\textbf{53.37} \\ \large $ \pm $0.04}                  & \makecell[c]{\vspace{-1mm}\textbf{\textcolor{red}{80.61}} \\ \large $ \pm $0.47} & \makecell[c]{\vspace{-1mm}\textbf{\textcolor{red}{92.34}} \\ \large $ \pm $0.64} & \makecell[c]{\vspace{-1mm}\textbf{\textcolor{red}{71.71}} \\ \large $ \pm $1.08} & \makecell[c]{\vspace{-1mm}\textbf{\textcolor{red}{84.70}} \\ \large $ \pm $1.26} & \makecell[c]{\vspace{-1mm}\textbf{\textcolor{red}{72.43}} \\ \large $ \pm $1.26} & \makecell[c]{\vspace{-1mm}\textbf{85.13} \\ \large $ \pm $1.14}                  & \makecell[c]{\vspace{-1mm}\textbf{\textcolor{red}{80.54}} \\ \large $ \pm $0.50} & \makecell[c]{\vspace{-1mm}\textbf{\textcolor{red}{92.35}}\\ \large $ \pm $0.64} &   \makecell[c]{\vspace{-1mm}\textbf{\textcolor{red}{70.31}}\\ \large $ \pm $11.71} &   \makecell[c]{\vspace{-1mm}\textbf{\textcolor{red}{82.51}}\\ \large $ \pm $12.34}                                 \\
\hline
\end{tabular}
}
\end{table*}

\section{Experiments}
\subsection{Datasets}
We evaluate the proposed CMML framework on two public (Derm7pt~\citep{kawahara2018seven}, ODIR~\citep{odir2019}) and one in-house (MEN) datasets, with image samples shown in Figure~\ref{data_show}.

\subsubsection{Derm7pt Dataset}
The skin lesion dataset contains 1011 cases of paired clinical and dermatoscopic images with five clinical attributes: sex, management, lesion location, lesion elevation, and level of diagnostic difficulty. Following~\citep{patricio2023coherent,hou2024concept}, we selected 827 cases diagnosed with melanoma (MEL) or nevus (NEV) diagnoses. All images were resized to 224$\times$224.

\subsubsection{ODIR Dataset}
The ocular disease dataset contains 3500 cases featuring paired left and right fundus images and two clinical attributes: sex and age. We concentrate on predicting disease categories, specifically Normal, Diabetes and Other diseases. Cases with multiple diseases were excluded, yielding a final subset of 2,641 cases. All images were resized to 512$\times$512.

\subsubsection{MEN Dataset}
The meningioma dataset was collected from the Brain Medical Center of Tianjin University, Tianjin Huanhu Hospital\footnote{This study was approved by the Institutional Review Board of Tianjin Huanhu Hospital (Jinhuan Ethical Review No. 2022-046), and the requirement for informed consent was waived.}.
This in-house dataset comprises 796 cases, featuring three MRI sequences: Contrast-Enhanced T1 (T1C), Contrast-Enhanced T2 FLAIR (FLAIR-C), and Apparent Diffusion Coefficient (ADC), along with six clinical attributes: sex, age, tumor area, edema area, tumor location, and the value of the apparent diffusion coefficient.
These cases are categorized into three grades: Grade 1 (G1, 650), Grade 2 with invasion (G2inv, 60), and Grade 2 without invasion (G2ninv, 86). Following~\citep{liu2025completed}, the regions of interest (ROIs) encompassing the tumor and edema were zero-padded to square shapes and resized to 24$\times$128$\times$128.



\subsection{Experimental Settings}
All models are trained for $50$ epochs using the Adam optimizer with a weight decay of $1 \times 10^{-4}$ and a $0.5$ dropout rate to prevent overfitting. A linear learning rate warm-up is applied during the first 5 epochs, where modality dropout is disabled to ensure stable populating of the memory banks. Subsequently, the learning rate follows a step-decay schedule, decreasing by $20\%$ every $5$ epochs.
Following~\citep{liu2026cfcml}, tabular attributes are transformed into textual descriptions via the templates in Table~\ref{template}, with the DeepSeek-R1 dimension $d_r=1536$ and the unified dimension $d=256$. The number of context tokens $N_{ct}$ is set to the average token count across all modalities. A consistent head count of four is maintained for all multi-head attention mechanisms across the CRTA and MHCA layer.
For the Derm7pt dataset, the model is trained with three random seeds at an initial learning rate of $5 \times 10^{-4}$ and a batch size of $64$. Following~\citep{tang2022fusionm4net}, we employ a Swin Transformer~\citep{liu2021swin} pre-trained on ImageNet-1k as the image encoder, with data augmentation including random shifts,  horizontal and vertical flips, and distortions.
For the ODIR and MEN datasets, performance is evaluated via 3-fold cross-validation. ODIR uses an initial learning rate of $1 \times 10^{-4}$ and a batch size of $90$, utilizing the frozen visual encoder from the FLAIR model~\citep{silva2025foundation} as the image backbone. 
MEN is trained with an initial learning rate of $5 \times 10^{-4}$ and a batch size of $42$, using nnMamba~\citep{gong2025nnmamba} as the image encoder.
Data augmentation for ODIR includes random horizontal flips, affine transformations and color jitter, while MEN employs random crop, random flip, gaussian noise, and random erasing~\citep{zhong2020random}.
Regarding the objective function, as individual loss components exhibit comparable numerical scales, all balancing hyperparameters ($\alpha, \beta, \gamma$) are set to unity. The memory update epoch $N_e$ is set to $25$ for all datasets, while the memory token count $N_t$ is configured to $128$ for the Derm7pt and MEN datasets, and $512$ for the ODIR dataset. 
All experiments were implemented in PyTorch and conducted on an NVIDIA GeForce RTX 3090 GPU.

\begin{figure}
    \centering
    \includegraphics[width=\linewidth]{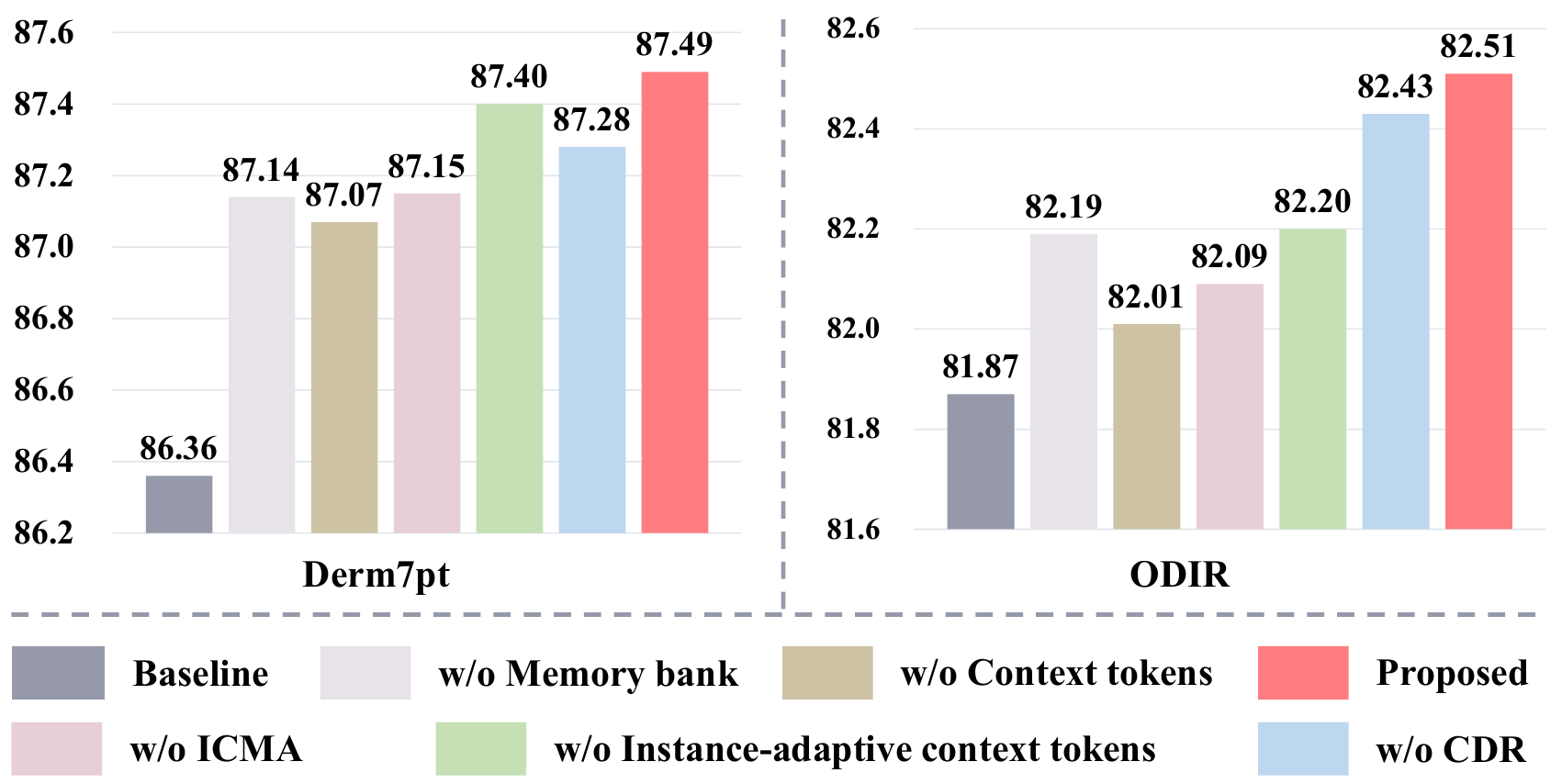}
    \caption{Ablation results for various components in proposed framework on both adopted datasets, showing the AVG AUC metric.
    }
    \label{ablation}
\end{figure}
\subsection{Quantitative Results}
We evaluated the proposed CMML framework against several SOTA methods, including completion-based approaches (Redcore~\citep{sun2024redcore}, MCMoE~\citep{xu2026mcmoe}) and joint representation-based methods (ShaSpec~\citep{wang2023multi}, SFusion~\citep{liu2023sfusion}, and SimMLM~\citep{li2025simmlm}). For a fair comparison, all methods utilized the same backbone encoders and were evaluated using Accuracy (ACC) and Area Under the Curve (AUC). 
We also performed Wilcoxon signed-rank tests~\citep{wilcoxon1992individual} to evaluate the significance of performance gains over existing SOTA methods.

The comprehensive results across various missing-modality scenarios for the Derm7pt and ODIR datasets are summarized in Table~\ref{result_comparison}. As illustrated, our CMML framework consistently achieves superior performance. On the Derm7pt dataset, it yields 80.57\% AVG ACC and 87.49\% AVG AUC, outperforming the optimal comparison method, SFusion, by 1.67\% and 1.69\%, respectively. Statistical tests confirm that these improvements are significant (p-value$<$0.05). While SFusion demonstrates strong performance by focusing on available modalities, our framework further enhances results by integrating synthesized representations for missing modalities. Specifically, compared to the completion-based MCMoE, CMML achieves significant improvements of 2.30\% in AVG ACC and 1.26\% in AVG AUC on Derm7pt, and 2.54\% and 0.97\% on ODIR.
This performance is driven by the CRTA's capture of inter-modal dependencies facilitated by integrated context tokens, complemented by the memory bank’s modality-specific enrichment.

\subsection{Ablation Studies}
To validate the effectiveness of the proposed framework, we conducted ablation experiments with both CMC and ICAR modules on the Derm7pt and ODIR datasets, as illustrated in Figure~\ref{ablation}. Specifically, the `Baseline' item denotes the removal of both CMC and ICAR from the proposed framework. The comparison between the `Baseline' and `Proposed' items indicates that the removal of these modules results in drops of $1.13\%$ and $0.64\%$ in the AVG AUC metric on the Derm7pt and ODIR datasets, respectively. 

\begin{figure}
    \centering
    \includegraphics[width=\linewidth]{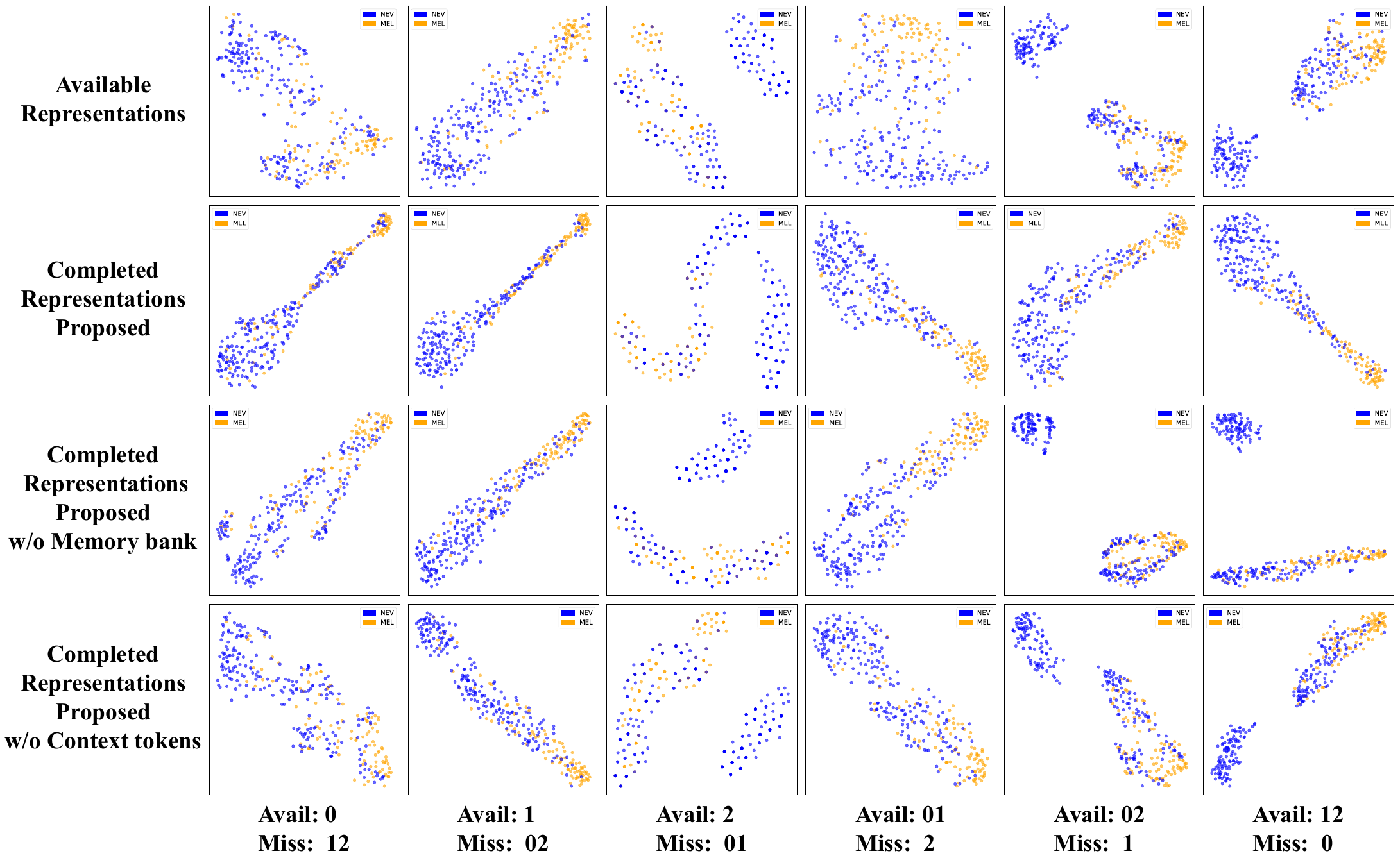}
    \caption{t-SNE visualization for available  and completed representations on the Derm7pt dataset under arbitrary missing scenarios. Columns denote modality availability patterns (0: clinical image, 1: dermatoscopic image, 2: tabular data). Points are colored by diagnostic class. Multimodal features are concatenated and GAP-processed for visualization.
    }
    \label{tsne_avail_comp}
\end{figure}
\begin{figure*}[t]
    \centering
    \includegraphics[width=\linewidth]{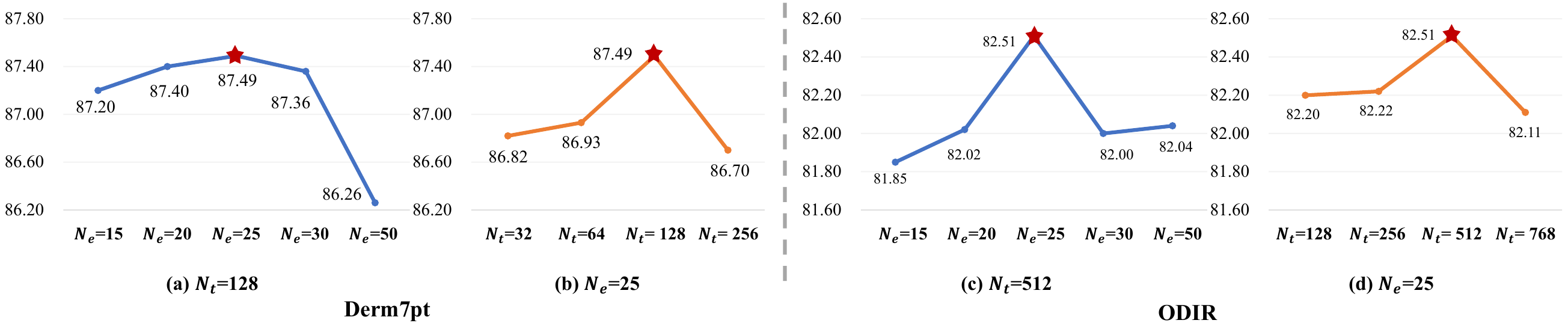}
    \caption{Display of the ablation study results regarding various memory update epochs and memory token lengths.}
    \label{ablation_memory}
\end{figure*}

\subsubsection{Effectiveness of the CMC module}
We performed two ablation studies for the CMC module focusing on the memory bank and context tokens, where ablation results are summarized in Figure~\ref{ablation}. On the Derm7pt dataset, removing the memory bank (w/o Memory bank) leads to a performance decline of 0.35\% in AVG AUC, while the exclusion of context tokens\footnote{The context token is utilized in both the CMC and ECAR modules; therefore, the removal of the context token renders the ECMA in the ECAR.} (w/o Context tokens) results in a 0.42\% reduction. Similarly, for the ODIR dataset, these two configurations lead to a decline of 0.32\% and 0.50\%, respectively, indicating a consistent performance drop across different missing scenarios when these components are absent.

To further evaluate the quality of the synthesized features, we visualize the class distributions of both available and completed representations using t-SNE~\citep{van2008visualizing} (Figure~\ref{tsne_avail_comp}). The results demonstrate a stark contrast between the available (first row) and completed representations (second row). The completed representations exhibit significantly clearer inter-class separation and enhanced intra-class compactness. Notably, the full proposed configuration (second row) visually outperforms the ablated variants without the memory bank (third row) and without context tokens (last row).

These quantitative and qualitative results collectively underscore the critical synergy between both components. The context tokens play a foundational role in capturing dataset-level inter-modal dependencies and providing prior guidance for synthesis, while the memory bank effectively supplements synthesized features with modality-specific information. Together, they enable the framework to recover essential semantic information and produce high-quality representations that are both robust and discriminative for missing-modality diagnosis.

\begin{figure}
    \centering
    \includegraphics[width=0.9\linewidth]{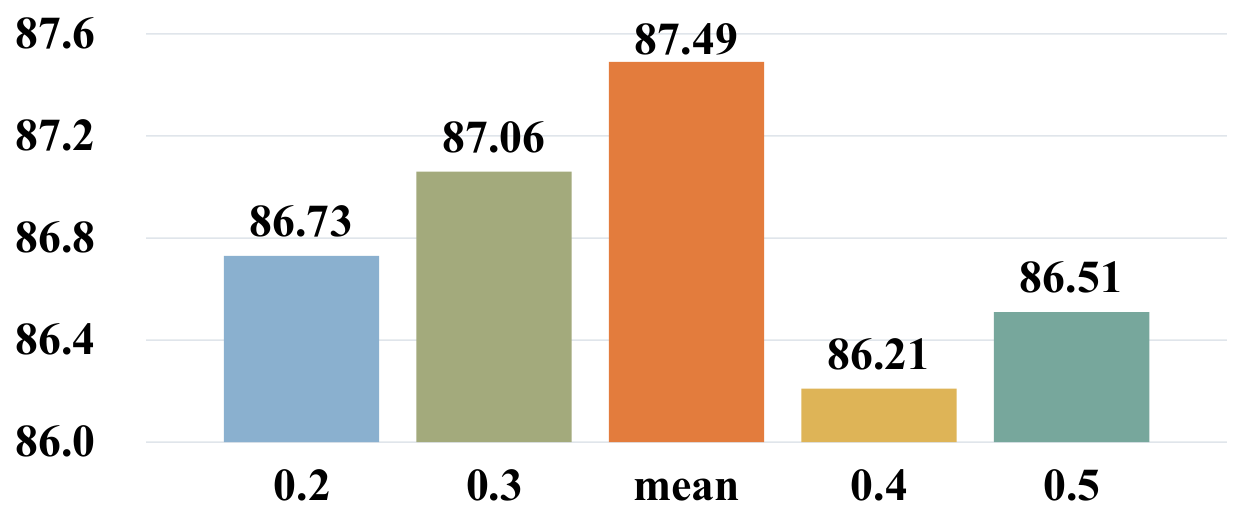}
    \caption{Ablation results for various ratios of context tokens to the total number of modality tokens in the Derm7pt dataset, illustrating the AVG AUC metric.
    }
    \label{ablation_ratio}
\end{figure}

\subsubsection{Effectiveness of the ICAR Module}
The efficacy of the ICMA component, instance-adaptive context tokens, and the CDR strategy is further validated through ablation studies (Figure~\ref{ablation}). Compared to the non-aligned baseline (w/o ICMA), integrating the ICMA module (Proposed) yields significant gains of 0.34\% and 0.42\% in AVG AUC for the Derm7pt and ODIR datasets, respectively. This improvement underscores the module's capacity in harmonizing the distributions between original available and synthesized representations, thereby effectively bridging the semantic modality gap.

Within the ICMA module, we further investigated the transition of context tokens from dataset-level priors to instance-adaptive references. Compared to the baseline (w/o ICMA), alignment guided solely by dataset-level semantic priors (w/o instance-adaptive context tokens) improves results by 0.25\% and 0.11\% for the Derm7pt and ODIR datasets, respectively. More importantly, alignment guided by instance-adaptive semantic reference (Proposed vs. w/o instance-adaptive context tokens) provides an additional boost of 0.09\% and 0.31\% on the respective datasets. This consistent progression confirms that infusing refined multimodal features into the context tokens creates a more representative instance-adaptive semantic reference, facilitating more precise, instance-driven alignment.

Finally, removing the CDR strategy (w/o CDR) results in a performance decline of 0.21\% on Derm7pt and 0.08\% on ODIR. This highlights the importance of class-aware contrastive learning in refining the integrated representations and promoting intra-class compactness, which ultimately bolsters the model's discriminative power for downstream diagnosis.

\begin{figure}
    \centering
    \includegraphics[width=\linewidth]{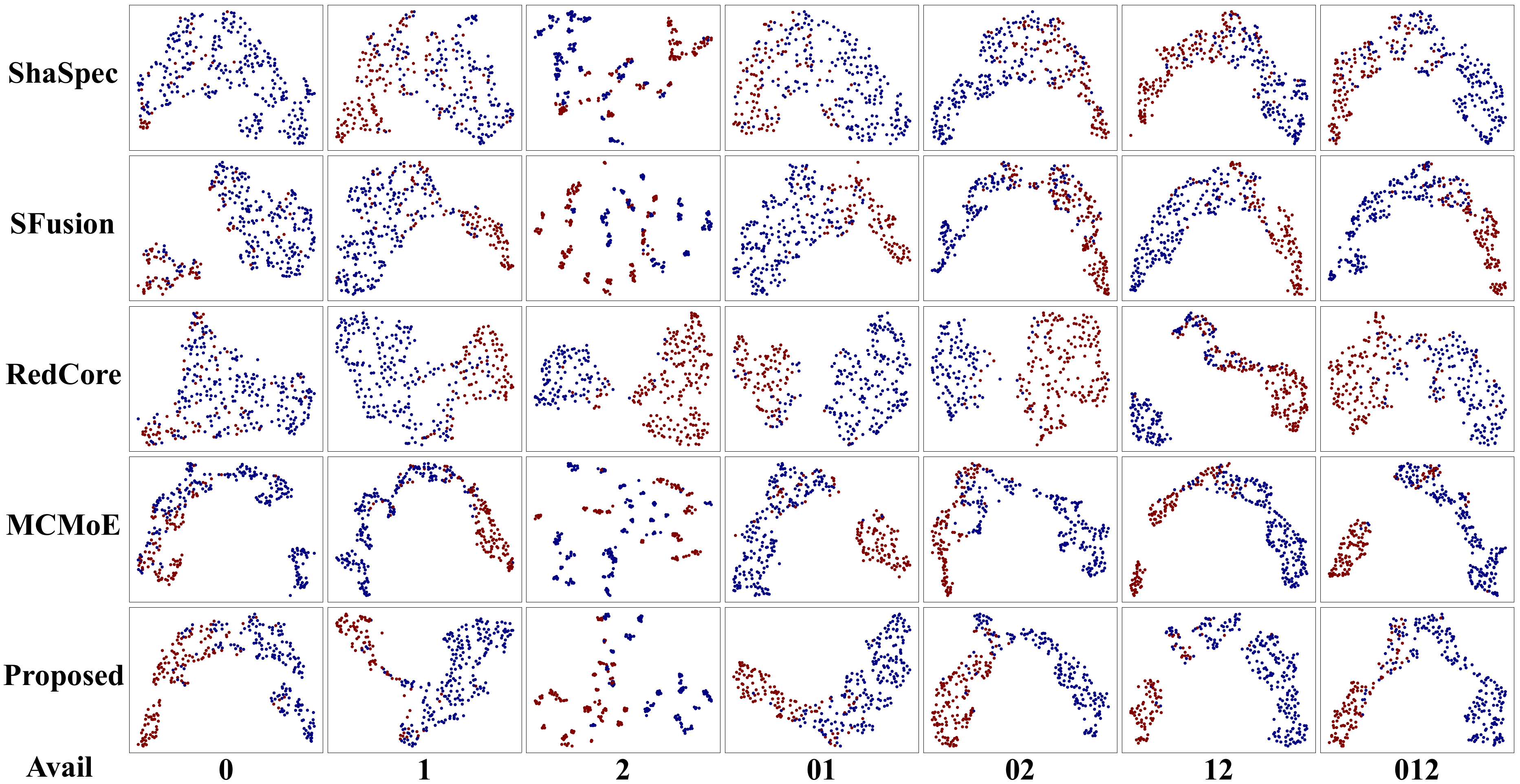}
    \caption{The visualization results of fused feature for the proposed and comparison methods utilizing the MDA method on the Derm7pt dataset. In each sub-figure, each point corresponds to a sample from the test set, with different colors indicating various classes.}
    \label{mda}
\end{figure}
\subsection{Parameter Sensitivity Analysis}
\subsubsection{Analysis of the memory bank}
The performance of the memory bank is governed by the update epoch $N_e$ and token length $N_t$. To determine the optimal configuration, we adopted a sequential optimization strategy (Figure~\ref{ablation_memory}). For Derm7pt, we first varied $N_e\in\{15, 20, 25, 30, 50\}$ with $N_t$ fixed at 128, observing peak performance at $N_e=25$. Subsequently, with $N_e$ fixed at 25, we evaluated $N_t\in\{32, 64, 128, 256\}$ and identified 128 as the optimal token length. Following the same methodology for ODIR, we evaluated $N_t\in\{128, 256, 512,768\}$ and established the final parameters as $N_e=25$ and $N_t=512$.

As $N_e$ increases, the memory bank's ability to capture discriminative modality-specific information improves. However, the performance decline observed beyond 25 epochs across both datasets suggests that excessive memory updates can lead to the accumulation of modality-specific noise or overfitting to outliers, which ultimately compromises the quality of the synthesized representations.

\begin{table*}
\centering
\caption{The comparison results on the MEN dataset. The notation in the "Available Modalities" is defined as follows, 0: T1C, 1: FLAIR-C, 2: ADC, 3: tabular data.}
\label{result_comparison_men}
\resizebox{\textwidth}{!}{
\Large
\begin{tabular}{c|ccccccccccccccc|c}
\hline
\multirow{2}{*}{Methods} & \multicolumn{15}{c|}{Available Modalities}                                                                                                                                                          & \multirow{2}{*}{\makecell[c]{AVG\\AUC}}  \\
\cline{2-16}
                         & 0           & 1           & 2          & 3          & 01           & 02          & 03          & 12         & 13         & 23         & 012         & 013         & 023         & 123        & 0123        &                          \\
\hline
SFusion~\cite{liu2023sfusion}                   & \makecell[c]{\vspace{-1mm} \textbf{\textcolor{red}{61.05}}\\ \large $ \pm $ 8.55}   & \makecell[c]{\vspace{-1mm} \textbf{64.25}\\ \large $ \pm $ 6.27}   & \makecell[c]{\vspace{-1mm} \textbf{95.86}*\\ \large $ \pm $ 4.35}  & \makecell[c]{\vspace{-1mm} \textbf{68.46}*\\ \large $ \pm $ 3.44}  & \makecell[c]{\vspace{-1mm} \textbf{\textcolor{red}{65.63}}\\ \large $ \pm $ 8.77}   & \makecell[c]{\vspace{-1mm} \textbf{95.44}* \\ \large $ \pm $ 3.13}  & \makecell[c]{\vspace{-1mm} \textbf{\textcolor{red}{72.04}}\\ \large $ \pm $ 5.53}  & \makecell[c]{\vspace{-1mm} \textbf{95.52}* \\ \large $ \pm $ 3.59}  & \makecell[c]{\vspace{-1mm} \textbf{71.71}* \\ \large $ \pm $ 2.79}  & \makecell[c]{\vspace{-1mm} \textbf{95.82}* \\ \large $ \pm $ 1.86}  & \makecell[c]{\vspace{-1mm} \textbf{95.57}* \\ \large $ \pm $ 3.24}  & \makecell[c]{\vspace{-1mm} \textbf{\textcolor{red}{73.52}}\\ \large $ \pm $ 6.36}  & \makecell[c]{\vspace{-1mm} \textbf{95.71}* \\ \large $ \pm $ 1.00}  & \makecell[c]{\vspace{-1mm} \textbf{95.67}* \\ \large $ \pm $ 1.37}  & \makecell[c]{\vspace{-1mm} \textbf{95.77}* \\ \large $ \pm $ 1.86}  & \makecell[c]{\vspace{-1mm} \textbf{82.80}* \\ \large $ \pm $ 13.29}               \\
RedCore~\cite{sun2024redcore}                   & \makecell[c]{\vspace{-1mm} 53.56* \\ \large $ \pm $ 16.08}  & \makecell[c]{\vspace{-1mm} 58.13* \\ \large $ \pm $ 11.83}  & \makecell[c]{\vspace{-1mm} 88.91*\\ \large $ \pm $ 5.35}  & \makecell[c]{\vspace{-1mm} 67.28* \\ \large $ \pm $ 3.37}  & \makecell[c]{\vspace{-1mm} 58.56* \\ \large $ \pm $ 10.86}  & \makecell[c]{\vspace{-1mm} 90.33* \\ \large $ \pm $ 5.41}  & \makecell[c]{\vspace{-1mm} 69.20\\ \large $ \pm $ 1.38}  & \makecell[c]{\vspace{-1mm} 88.91* \\ \large $ \pm $ 5.05}  & \makecell[c]{\vspace{-1mm} 68.84* \\ \large $ \pm $ 4.27}  & \makecell[c]{\vspace{-1mm} 87.80* \\ \large $ \pm $ 3.31}  & \makecell[c]{\vspace{-1mm} 89.32* \\ \large $ \pm $ 4.09}  & \makecell[c]{\vspace{-1mm} 69.72* \\ \large $ \pm $ 3.21}  & \makecell[c]{\vspace{-1mm} 88.50* \\ \large $ \pm $ 2.82}  & \makecell[c]{\vspace{-1mm} 88.95* \\ \large $ \pm $ 1.67}  & \makecell[c]{\vspace{-1mm} 89.56* \\ \large $ \pm $ 0.86}  & \makecell[c]{\vspace{-1mm} 77.17* \\ \large $ \pm $ 11.96}              \\
MCMoE~\cite{xu2026mcmoe}                        & \makecell[c]{\vspace{-1mm} 56.57* \\ \large $ \pm $ 7.04}   & \makecell[c]{\vspace{-1mm} 63.19* \\ \large $ \pm $ 11.96}  & \makecell[c]{\vspace{-1mm} 93.77* \\ \large $ \pm $ 2.30}  & \makecell[c]{\vspace{-1mm} 66.67* \\ \large $ \pm $ 9.12}  & \makecell[c]{\vspace{-1mm} 62.96* \\ \large $ \pm $ 14.07}  & \makecell[c]{\vspace{-1mm} 93.55* \\ \large $ \pm $ 2.05}  & \makecell[c]{\vspace{-1mm} 67.03* \\ \large $ \pm $ 9.07}  & \makecell[c]{\vspace{-1mm} 93.54* \\ \large $ \pm $ 1.90}  & \makecell[c]{\vspace{-1mm} 68.50* \\ \large $ \pm $ 9.02}  & \makecell[c]{\vspace{-1mm} 88.49* \\ \large $ \pm $ 4.47}  & \makecell[c]{\vspace{-1mm} 93.16* \\ \large $ \pm $ 2.35}  & \makecell[c]{\vspace{-1mm} 68.32* \\ \large $ \pm $ 9.05}  & \makecell[c]{\vspace{-1mm} 88.24* \\ \large $ \pm $ 5.21}  & \makecell[c]{\vspace{-1mm} 88.87* \\ \large $ \pm $ 4.20}  & \makecell[c]{\vspace{-1mm} 88.26* \\ \large $ \pm $ 5.03}  & \makecell[c]{\vspace{-1mm} 78.74* \\ \large $ \pm $ 16.39}              \\
\hline
Ours                      & \makecell[c]{\vspace{-1mm} \textbf{56.73}\\ \large $ \pm $ 10.87}  & \makecell[c]{\vspace{-1mm} \textbf{\textcolor{red}{65.25}}\\ \large $ \pm $ 3.67}   & \makecell[c]{\vspace{-1mm} \textbf{\textcolor{red}{97.63}}\\ \large $ \pm $ 1.99}  & \makecell[c]{\vspace{-1mm} \textbf{\textcolor{red}{73.14}}\\ \large $ \pm $ 3.67}  & \makecell[c]{\vspace{-1mm} \textbf{65.14}\\ \large $ \pm $ 1.74}   & \makecell[c]{\vspace{-1mm} \textbf{\textcolor{red}{97.80}}\\ \large $ \pm $ 1.70}  & \makecell[c]{\vspace{-1mm} \textbf{71.43}\\ \large $ \pm $ 1.92}  & \makecell[c]{\vspace{-1mm} \textbf{\textcolor{red}{97.92}}\\ \large $ \pm $ 1.56}  & \makecell[c]{\vspace{-1mm} \textbf{\textcolor{red}{73.99}}\\ \large $ \pm $ 2.42}  & \makecell[c]{\vspace{-1mm} \textbf{\textcolor{red}{97.99}}\\ \large $ \pm $ 2.55}  & \makecell[c]{\vspace{-1mm} \textbf{\textcolor{red}{97.92}}\\ \large $ \pm $ 1.02}  & \makecell[c]{\vspace{-1mm} \textbf{72.77}\\ \large $ \pm $ 1.54}  & \makecell[c]{\vspace{-1mm} \textbf{\textcolor{red}{98.05}}\\ \large $ \pm $ 3.04}  & \makecell[c]{\vspace{-1mm} \textbf{\textcolor{red}{98.07}}\\ \large $ \pm $ 2.03}  & \makecell[c]{\vspace{-1mm} \textbf{\textcolor{red}{98.00}}\\ \large $ \pm $ 1.56}  & \makecell[c]{\vspace{-1mm} \textbf{\textcolor{red}{84.12}} \\ \large $ \pm $ 13.22}             \\
\hline
\end{tabular}}
\end{table*}

\subsubsection{Impact of Context Token Quantity}
We evaluate the impact of context token quantity on Derm7pt by comparing fixed ratios $\rho \in \{0.2,0.3,0.4,0.5\}$ (relative to the total modality tokens) against an adaptive ``mean" setting, where the token count equals the average length of all modalities. As shown in Figure~\ref{ablation_ratio}, the ``mean" configuration achieves the best performance. This suggests that a context capacity matching the average information density of modalities provides more robust dataset-level semantic priors for capturing inter-modal dependencies than heuristic fixed-ratio settings.

\subsection{Feature Distribution Visualization}
Using Manifold Discovery and Analysis (MDA)\citep{islam2023revealing}, we visualized the latent space of fused multimodal features for the Derm7pt dataset (Figure\ref{mda}). Our CMML framework consistently exhibits tighter intra-class clustering and superior inter-class separation compared to SOTA methods. Particularly in scenarios where only clinical images are available, while competing methods fail to establish clear decision boundaries, our method maintains high discriminative power. This advantage is attributed to the CMC module's ability to recover essential semantic features and the ICAR module's efficacy in aligning and refining diagnostic cues through a shared semantic reference.

\subsection{Scalability to Scenarios with Increased Modalities}
To evaluate the scalability of our framework, we conducted experiments on the four-modal MEN dataset, which presents a more complex environment for capturing inter-modal dependencies. As shown in Table~\ref{result_comparison_men}, CMML achieves the highest AVG AUC of 84.12\% compared to the other methods\footnote{The ShaSpec and SimMLM are excluded because they encounter significant challenges in this dataset, struggling to provide effective classification.}. These results underscore the robust scalability of the proposed CMML framework, demonstrating its flexibility to adapt to varying modality counts without the need for complex, modality-specific architectural adjustments. This adaptability is crucial for real-world clinical applications where the number and types of imaging sequences can vary significantly.

\begin{table}
\centering
\caption{Evaluation of parameter 
and computational complexity 
on the adopted datasets. }
\label{computational_complexity}
\Large
\renewcommand{\arraystretch}{1}
\resizebox{0.45\textwidth}{!}{
\begin{tabular}{c|cc|cc|cc} 
\hline
\multirow{2}{*}{Methods} & \multicolumn{2}{c|}{Derm7pt} & \multicolumn{2}{c|}{ODIR} & \multicolumn{2}{c}{MEN}  \\ 
\cline{2-7}
                         & Param & GFLOPs               & Param & GFLOPs            & Param & GFLOPs           \\ 
\hline
ShaSpec~\cite{wang2023multi}                  & 28.47 & 8.76                 & 1.61  & 43.44             & -     & -                \\
SFusion~\cite{liu2023sfusion}                  & 34.93 & 8.79                 & 8.11  & 43.47             & 17.41 & 6.97             \\
RedCore~\cite{sun2024redcore}                  & 41.81 & 8.94                 & 14.99 & 43.48             & 32.16 & 7.13             \\
SimMLM~\cite{li2025simmlm}                   & 83.87 & 8.74                 & 2.30   & 43.18             & -     & -                \\
MCMoE~\cite{xu2026mcmoe}                    & 35.65 & 8.94                 & 8.82  & 43.63             & 19.83 & 7.21             \\ 
\hline
Proposed                 & 30.71 & 8.96                 & 4.27  & 44.15             & 14.00    & 7.55             \\
\hline
\end{tabular}}
\end{table}


\subsection{Computational Complexity Analysis}
As summarized in Table~\ref{computational_complexity}, CMML offers a favorable trade-off between diagnostic accuracy and computational cost. By maintaining a leaner parameter count and comparable GFLOPs to leading comparison methods, our framework demonstrates high efficiency, supporting its potential for deployment in resource-constrained clinical environments.

\section{Discussion and Conclusion}
In this paper, we proposed CMML, a innovative framework for robust multimodal diagnosis under arbitrary missing scenarios via a “synthesis-then-alignment” strategy. Its CMC module leverages the CRTA and context tokens to synthesize base representations, which are further enriched with modality-specific information from memory banks. To bridge the semantic gap between original available and synthesized representations, we transform the learned context priors into instance-adaptive semantic references for feature alignment, followed by class-aware contrastive refinement to promote discriminability. Evaluations on three diverse datasets demonstrate that CMML significantly outperforms SOTA methods by recovering essential missing information and ensuring multimodal consistency across heterogeneous data.

The core technical strength of CMML lies in the dynamic evolution of context tokens. During the synthesis phase, these tokens function as dataset-level semantic prior to capture inter-modal dependencies. We transform them into instance-adaptive semantic references that ensure that the alignment process is meticulously tailored to the unique pathological profile of each patient, effectively bridging the semantic modality gap while preserving critical diagnostic cues that are essential for accurate reasoning. Beyond precision, this hub-centric architecture is also fundamental to the framework's robust scalability. By routing all inter-modal interactions through a unified context space, CMML effectively manages the combinatorial complexity of missing patterns without requiring structural reconfiguration,  as evidenced by its superior performance on three- and four-modality datasets..

Despite its efficacy, CMML has limitations. First, its dependency on complete training sets restricts the utilization of vast, partially complete clinical registries. Second, modality imbalance persists; synthesized features may fail to capture the critical details needed to replace a missing primary modality. Future work will explore self-supervised pre-training and importance-aware loss weighting to address these limitations.
\printcredits

\bibliographystyle{cas-model2-names}

\bibliography{ref}



\end{document}